\documentclass[letterpaper, 10 pt, conference]{ieeeconf} 

\IEEEoverridecommandlockouts                              % This command is only needed if 
\usepackage{cite}
\usepackage{amsmath,amssymb,amsfonts}
\usepackage{algorithmic}
\usepackage{graphicx}
\usepackage{textcomp}
\usepackage{xcolor}
\usepackage{color, colortbl}
\usepackage{multirow}
\usepackage{siunitx}
\usepackage{subcaption}
\usepackage{float}
\usepackage{stfloats}
\usepackage{hyperref}
\usepackage{fancyhdr}
\usepackage{booktabs}
\usepackage{microtype}

\hypersetup{
    colorlinks=true,
    linkcolor=black,
    filecolor=magenta,      
    urlcolor=cyan,
    pdftitle={Overleaf Example},
    pdfpagemode=FullScreen,
    }

\title{\LARGE \bf
MR3D-Net: Dynamic Multi-Resolution 3D Sparse Voxel Grid Fusion for LiDAR-Based Collective Perception
}

\author{Sven Teufel$^{1}$, Jörg Gamerdinger$^{1}$, Georg Volk$^{1}$ and Oliver Bringmann$^{1}$% <-this % stops a space
\thanks{$^{1}$University of Tübingen, Faculty of Science, Department of Computer Science, Embedded Systems {\tt\small \{sven.teufel, joerg.gamerdinger, georg.volk, oliver.bringmann\} @uni-tuebingen.de}}%
}

\begin{document}

\maketitle
\thispagestyle{empty}
\pagestyle{empty}
%%%%%%%%%%%%%%%%%%%%%%%%%%%%%%%%%%%%%%%%%%%%%%%%%%%%%%%%%%%%%%%%%%%%%%%%%%%%%%%%
\begin{abstract}
The safe operation of automated vehicles depends on their ability to perceive the environment comprehensively. However, occlusion, sensor range, and environmental factors limit their perception capabilities. To overcome these limitations, collective perception enables vehicles to exchange information. However, fusing this exchanged information is a challenging task. Early fusion approaches require large amounts of bandwidth, while intermediate fusion approaches face interchangeability issues. Late fusion of shared detections is currently the only feasible approach. However, it often results in inferior performance due to information loss. 
To address this issue, we propose MR3D-Net, a dynamic multi-resolution 3D sparse voxel grid fusion backbone architecture for LiDAR-based collective perception.
We show that sparse voxel grids at varying resolutions provide a meaningful and compact environment representation that can adapt to the communication bandwidth. 
MR3D-Net achieves state-of-the-art performance on the OPV2V 3D object detection benchmark while reducing the required bandwidth by up to 94\,\% compared to early fusion.
Code is available at \url{https://github.com/ekut-es/MR3D-Net}

\end{abstract}
%%%%%%%%%%%%%%%%%%%%%%%%%%%%%%%%%%%%%%%%%%%%%%%%%%%%%%%%%%%%%%%%%%%%%%%%%%%%%%%%
\section{Introduction}
\label{sec:intro}
Automated vehicles (AVs) require a comprehensive perception of their environment in order to operate safely in complex driving scenarios.
However, in real-world scenarios, AVs encounter limitations in their environmental perception due to infrastructural obstacles and limited sensing ranges~\cite{volk_environment-aware_2019}. In addition, the perception capabilities of AVs are negatively affected by adverse weather conditions such as rain, fog or snow~\cite{teufel2022simulating, teufel2023enhancing}.

A promising approach to overcome these problems is Collective Perception (CP), where information about the perceived environment are shared between different connected and automated vehicles (CAVs).
In order to build a meaningful local environment model, CP must address the challenge of fusing the locally perceived information with the collectively received information. There are several approaches to this fusion, which can be classified into three classes based on the type of data exchanged between CAVs: early, late, and intermediate fusion. 

For early fusion, CAVs share raw sensor data, which provides the highest information value since it includes all available information. Although early fusion is considered the most accurate approach due to the absence of information loss, it is not suitable for real-world applications because of bandwidth limitations in communication~\cite{yuan2022keypoints}.

Late fusion refers to the use of pre-processed information about detected objects. The CAVs exchange information regarding the position, velocity, and acceleration of detected objects in their surroundings.
This approach requires the lowest bandwidth~\cite{caillot2022survey} and can rely on the standardized Collective Perception Message (CPM)~\cite{CPM} which defines the exchange of detected objects. However, this fusion method is less accurate compared to early fusion due to information loss from pre-processing~\cite{caillot2022survey}.

Intermediate fusion aims to combine the benefits of early and late fusion by using features from the local object detector as exchanged information.
Considering LiDAR-based object detectors, possible features could be voxel features \cite{chen2019f} or point features \cite{yuan2022keypoints}.
Using these features reduces bandwidth compared to raw data in early fusion while maintaining higher accuracy than pre-processed data in late fusion. However, the application of intermediate fusion in the real world is limited, as all CAVs require identical detection architectures.

For real-world applications, a standardizable environment representation that can dynamically adapt to the communication channel load could significantly improve both the communication channel utilization and the perception performance.

Therefore, we propose \textbf{MR3D-Net}, a dynamic fusion approach that achieves state-of-the-art performance on the OPV2V benchmark while drastically reducing the bandwidth requirement compared to early fusion methods using a unified, adaptive, and exchangeable environment representation.
\ \\
\ \\[-0.3cm]
Our main contributions are:\\[-0.3cm]
\begin{itemize}
    \item We provide a dynamic environment representation utilizing 3D sparse voxel grids that can adapt the resolution based on the communication channel load.\\[-0.3cm]
    \item We present a novel and effective fusion architecture which enables the fusion of environmental information represented with multi-resolution sparse voxel grids.\\[-0.3cm]
    \item MR3D-Net achieves state of the art results on the OPV2V \cite{xu2022opv2v} 3D object detection benchmark.\\[-0.3cm]
\end{itemize}
\ \\
\ \\[-0.7cm]
In Section~\ref{rel_work} we introduce related work to our approach. Afterwards, we present a novel dynamic environment representation for our multi-resolution sparse voxel grid fusion pipeline in Sec.~\ref{sec:representation} and the corresponding fusion architecture in Sec.~\ref{sec:mr3dnet}. Section~\ref{sec:eval} describes the conducted experiments including the used dataset. We provide the results in Sec.~\ref{sec:results}. Finally, we give a conclusion and outlook on future research.

\newpage
\section{Related Work}
\label{rel_work}
\subsection{Object Detection}
There exists a variety of LiDAR-based object detectors. Yang et al. \cite{yang2018pixor} proposed PIXOR, which is an one-stage object detector that employs 2D convolutions on point clouds that are mapped into bird's-eye view (BEV) feature maps. 
VoxelNet by Zhou et al. \cite{zhou2018voxelnet} is a one-stage detector that introduced the concept of voxel feature encoders. This approach employs a PointNet \cite{qi2017pointnet} module to encode features from the points to a voxel grid, after which 3D convolutions are applied.  
SECOND by Yan et al. \cite{yan2018second}, is also a one-stage detector that employs 3D convolutions on voxelized point clouds. However, they implemented efficient 3D spatially sparse and submanifold convolutions to reduce the computational effort on sparse 3D point clouds. 
PointPillars by Lang et al. \cite{lang2019pointpillars} is a one-stage detector that introduced the concept of pillars, which are voxels that span the entire vertical axis. After applying a voxel feature encoder similar to SECOND, a 2D BEV feature map is obtained on which 2D convolutions are applied.
Shi et al. \cite{shi2023pv} proposed PV-RCNN++, a two-stage LiDAR-based object detector that employs both a voxel-based and a point-based representation. The point cloud is voxelized and then processed by a 3D sparse convolutional backbone, as in SECOND, followed by a 2D Region Proposal Network (RPN) that generates 3D bounding box proposals. The second stage then samples keypoints, which are used to aggregate the sparse voxel features from the 3D CNN along with 2D BEV features from the RPN. Subsequently these keypoint features are pooled and used as input to the detection heads.

\subsection{Early Fusion}
For the early fusion approach, raw sensor data is exchanged between agents. Since only the point clouds of other agents are transformed into the ego coordinate system and then added to the local point cloud, all LiDAR-based object detectors can be used for early fusion. Chen et al. \cite{chen2019cooper} used early fusion on their T\&J dataset as well as on KITTI \cite{geiger2012we}. 
They could show a significant improvement in detection performance, however the KITTI dataset is not dedicated to collective perception since it was recorded with a single vehicle and the T\&J dataset only consists of a few frames on a parking lot.

The authors state, that transmitting raw point clouds with existing vehicular networks is feasible. However, they consider only the transmission of partial low resolution point clouds at a low frequency of \SI{1}{\hertz}.
Arnold et al.~\cite{arnold2020cooperative} investigated fusion strategies for collective perception with infrastructural LiDAR sensors. They could show, that the early fusion outperforms late fusion in the evaluated scenarios. However, they used stationary sensors, that can use wired communication, hence, it can not be compared to V2X communication.
Xu et al. \cite{xu2022opv2v} benchmarked several LiDAR-based object detectors on their OPV2V dataset using early, intermediate, and late fusion. Their results showed that late fusion performed the worst and intermediate fusion performed the best. However, in most cases, the intermediate fusion performed only slightly better than the early fusion.

\subsection{Late Fusion}
For late fusion pre-processed data in the form of detected bounding boxes is exchanged between vehicles. A simple way to realize late fusion is to use detected bounding boxes from multiple collaborative agents and weight them based on their detection confidence. Solovyev et. al~\cite{solovyev2021weighted} introduced a Weighted Mean Fusion (WMF) for local image-based detection.
Houenou et. al~\cite{houenou_track--track_2012} and Müller et. al~\cite{muller_generic_2011} applied WMF for multisensor fusion of vehicle-local sensor data. Aeberhard~\cite{aeberhard_object-level_2017} extended this approach to a WMF for collective perception.
Another way of late fusion is to use an adapted Kalman filter. The collectively perceived tracks are considered as measurements and integrated into the local environmental model. Approaches using Kalman filter based collective perception are presented in~\cite{allig2019,volk_environment-aware_2019,gabb2019infrastructure,volk2021}. 
Another late fusion method is to directly fuse the collective detections within the local perception pipeline using neural networks \cite{teufel2023collective}. This method could significantly outperform WMF, however there is still a substantial loss of information when only detections are shared.

\section{Environment Representation}
\label{sec:representation}

In order to realize collective perception, an exchangeable, compact, and standardizable representation of the environment is required to share information between vehicles. As stated in Section \ref{sec:intro}, exchanging raw sensor data is not feasible due to the high bandwidth requirement, and using intermediate fusion approaches is also unrealistic due to problems of interchangeability and non-standardization. Late fusion approaches are the only commonly used methods that satisfy the given constraints and can be deployed in real-world applications. However, they may suffer from lower performance since only the exchanged bounding boxes do not provide a comprehensive representation of the environment.  
Therefore, we propose exchanging sparse voxel grids between vehicles since they provide a geometric representation of the environment, significantly reducing the amount of transmitted data compared to exchanging raw data, and are interchangeable and standardizable. Table \ref{tab:msg_sizes} displays a comparison of message sizes between sparse voxel grids and raw data. Sparse voxel grids are a uniform partitioning of 3D space into equally sized voxels, where only the voxels containing data are stored. In our case, voxels are stored if they contain at least one LiDAR point. The sparse voxel grids are stored in coordinate format, i.e. for each non-empty voxel, the coordinates within the grid are stored. Typically, when the point cloud is voxelized, each voxel is assigned a feature vector derived from the points within the voxel. A common approach is to use the mean of each point feature for the points within a voxel\cite{shi2020pv}. As shown in Section \ref{sec:results}, we found that using the center of each voxel as the voxel feature achieved roughly the same performance as using the mean features, so we only transmit the coordinates of the voxels, as the center of each voxel can be inferred from the coordinates, voxel size and origin of the voxel grid.

\begin{table}
\centering
\caption{Comparison of raw point cloud and sparse voxel grid message sizes for the OPV2V \cite{xu2022opv2v} test set with $\approx$ 57,000 points on average per frame.}
\label{tab:msg_sizes}
\renewcommand{\arraystretch}{1.5}
\begin{tabular}{@{}lc@{}}
\toprule
Shared Data                                                                                     & Average Data Size [\si[per-mode=repeated-symbol]{\kilo\byte}]  \\ \midrule \midrule
Raw Point Clouds                                                                                & 914.9                      \\ 
\SI[parse-numbers = false]{\ \, 5\times \ \, 5\times10}{\centi\meter} Sparse Voxel Grids         & 180.0                      \\ 
\SI[parse-numbers = false]{10\times10\times20}{\centi\meter} Sparse Voxel Grids                  & 111.0                      \\ 
\SI[parse-numbers = false]{20\times20\times40}{\centi\meter} Sparse Voxel Grids                  & 54.5                       \\ 
\bottomrule
\end{tabular}
\end{table}

\section{MR3D-Net}
\label{sec:mr3dnet}
\begin{figure*}
    \centering
    \includegraphics[trim= 1.45cm 3cm 1.45cm 3cm, clip, width=\textwidth]{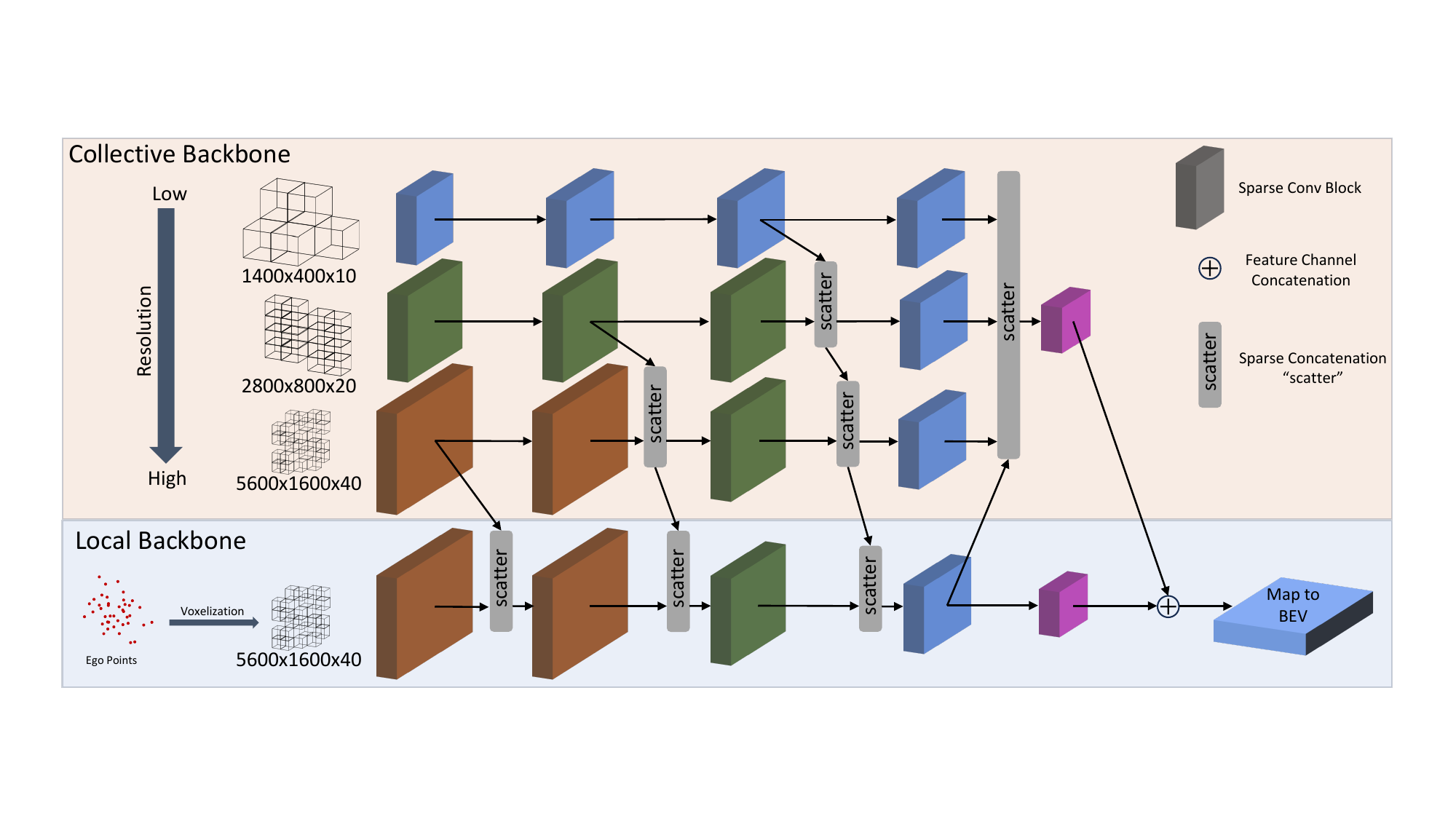}
    \caption{Architecture of MR3D-Net. The collective backbone consists of three input streams which take sparse voxel grids at different resolutions as input. The local backbone takes the voxelized ego point cloud as input and shares the structure with the high-resolution input stream from the collective backbone. The color of the sparse convolution blocks corresponds to their input resolution.}
    \label{fig:architecture_overview}
\end{figure*}

We propose a novel fusion approach to improve LiDAR-based collective perception by exchanging sparse voxel grids instead of raw sensor data, to heavily reduce the amount of data that needs to be transmitted. The advantage of our method is that the shared information is compact and therefore does not exceed the bandwidth limitations in communication. Furthermore, the shared information provides a meaningful representation of the environment, allowing for fusion using a variety of approaches and enabling it to be standardized. We demonstrate that this environment representation can effectively improve LiDAR-based collective perception. Therefore, we propose \textbf{MR3D-Net}, a multi-resolution sparse voxel grid fusion architecture that can be used as a backbone in various object detectors. 
An overview of the proposed MR3D-Net architecture is given in Figure \ref{fig:architecture_overview}.
We implemented MR3D-Net into the OpenPCDet\cite{openpcdet2020} toolbox for LiDAR-based 3D object detection.

\subsection{Architecture}
In order to dynamically adapt the communication to the current channel load, we leverage three input streams at different voxel grid resolutions. Each input stream consists of four consecutive convolution blocks, where a convolution block consists of a sparse convolution layer followed by two sub-manifold convolutions. After each of these layers, a batchnorm and a relu activation function are applied. To reduce the spatial dimensions of the voxel grid, the sparse convolution layer may be strided. The module containing these three input streams is called the collective backbone. Besides the collective backbone, MR3D-Net also contains a local backbone which takes the LiDAR points of the ego vehicle as input. The local backbone shares the same structure with the highest resolution input stream from the collective backbone. In order to fuse the information between the different resolution streams, we use the scatter operation.

\begin{figure}
    \centering
    \includegraphics[width=0.8\linewidth]{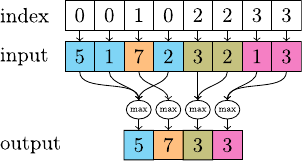}
    \caption{scatter operation in the 1-dimensional case using the $max$ function \cite{fey2019fast}}
    \label{fig:scatter}
\end{figure}

\subsubsection{Scatter Operation}
A simple feature concatenation of two voxel grids at the same resolution is not possible due to their sparsity, since they do not necessarily share the same voxels. Therefore, some voxels appear only in one of the two tensors, while others are contained in both. To still fuse the voxel features, we use a sparse update operation called scatter. The scatter operation takes both sparse voxel grids and concatenates them. Then, a permutation invariant function with arbitrarily many inputs is applied to duplicate voxels. Figure \ref{fig:scatter} shows an example in the one-dimensional case where the $max$ function is applied to duplicate indices. The resulting output is the union of the two tensors where the value of each voxel is determined by the applied function. Besides the $max$ function, other common functions are $min$, $sum$, $mean$, and $mul$.

The scatter operation is applied in MR3D-NET after the sparse convolution blocks with the $max$ function, where two input streams share the same spatial dimension. The output of the scatter operation is then used as input for the higher resolution stream. For the lower resolution stream, the output of the previous convolution block is used as input. This way, the different input streams can learn from their own input resolution as well as from the resolution below, enabling the neural network to extract features from multiple resolutions differently. After the four convolution blocks, all outputs of the collective backbone streams are then scattered with the output from the local backbone, and a final sparse convolution block is applied in both the collective and local backbone. The outputs of these convolutions then share the same voxels. However, the voxel features are different. Therefore, to obtain the final convolution output, we concatenate the features of both outputs. After that, we map the final voxel grid to bird's eye view by concatenating the voxel features along the z-axis, resulting in a 2D feature map that is then used by the object detector.

\subsubsection{Implementation Details}
For the three input streams, we selected the voxel sizes to be $\SI{5}{\centi\meter} \times \SI{5}{\centi\meter} \times \SI{10}{\centi\meter}$ for the high resolution input stream and the local backbone, for the medium resolution, we selected twice this size which is $\SI{10}{\centi\meter} \times \SI{10}{\centi\meter} \times \SI{20}{\centi\meter}$. The voxel size of the low resolution stream is four times the high resolution voxel size which is $\SI{20}{\centi\meter} \times \SI{20}{\centi\meter} \times \SI{40}{\centi\meter}$. For training and evaluation, we used a maximum grid size of $\SI{280}{\meter} \times \SI{80}{\meter} \times \SI{4}{\meter}$, resulting in input resolutions of $\SI{5600}{} \times \SI{1600}{} \times \SI{40}{}$, $\SI{2800}{} \times \SI{800}{} \times \SI{20}{}$ and $\SI{1400}{} \times \SI{400}{} \times \SI{10}{}$ voxels for the high, medium, and low input stream respectively. In the local backbone and the high-resolution input stream, the second, third, and fourth convolutions use a stride of two to reduce spatial dimensions. Since the input resolution of the medium-resolution stream is already half of the high input resolution, only the third and fourth convolutions use a stride of two. For the low resolution stream, the input is already half of the medium resolution and a quarter of the high resolution. Therefore, only the fourth convolution block uses a stride of two. The sparse convolution blocks at the different resolutions all share the same number of output channels, which are $[16, 32, 64, 64]$. The kernel size of all layers is $3\times3\times3$.
\section{Evaluation}
\label{sec:eval}
\vspace*{-0.3cm}
\subsection{Dataset}
For the evaluation of our method we used the OPV2V dataset \cite{xu2022opv2v}. OPV2V is a simulated dataset for collective perception which was generated using the OpenCDA \cite{xu2021opencda} co-simulation tool for CARLA \cite{CARLA} and SUMO \cite{SUMO2018}.  We trained the detectors on the training split which contains 6,765 samples, and evaluated them on the test set, consisting of 2,170 samples. One sample corresponds to one time step, i.e. multiple frames of different vehicles taken at the same time. For each sample, we randomly select which frame is the ego frame and use the other frames for the collectively shared information. The number of frames per sample, i.e. the number of CAVs in the scenario ranges between 2 and~7.
\subsection{Experiments}
For the evaluation of the proposed method, we used MR3D-Net as backbone for the PV-RCNN++ \cite{shi2023pv} object detector. We evaluated MR3D-Net with various resolution assignment strategies. First, we dynamically assigned the frames of the cooperative vehicles to the three input resolution streams of MR3D-Net using a uniform random distribution. Then, the point clouds are voxelized in the desired resolutions. We call this variant MR3D-Net Dynamic. Together with the ego point cloud these voxel grids are the only input to MR3D-Net, so only the sparse voxel grids need to be shared between vehicles. For the input streams where none of the cooperative frames was assigned, we used the voxelized ego point cloud at the respective resolution as input to speedup the training and to always learn from multiple input resolutions simultaneously.
Furthermore, to study the impact of different input resolutions on the detection performance, we evaluated each input resolution individually by assigning all cooperative frames only to one input stream, and using the voxelized ego point cloud at the respective resolution as input for the other two input streams.
To quantify the loss of information when only the voxel grid coordinates are transmitted instead of the grid with per voxel features, we also trained MR3D-Net with the mean of the points as voxel feature. 
As a baseline, we trained the vanilla PV-RCNN++ detector on the ego point clouds only, without any cooperative information. As a target value we also trained PV-RCNN++ in an early fusion manner with the ego point cloud and all point clouds from the cooperative vehicles as input.
We trained the MR3D-Net and PV-RCNN++ variants all for 120 epochs, for the remaining early fusion approaches we took the results from \cite{xu2022opv2v}.  
For the evaluated methods we report the Average Precision (AP) results on the OPV2V test set at an Intersection over Union (IoU) threshold of 0.7. As suggested by OPV2V \cite{xu2022opv2v}, we do not sort the detections by confidence for the AP calculation. We evaluate the detections within a range of $[-140, 140]$\,\si{\metre} in x-direction, $[-40, +40]$\,\si{\metre} in y-direction and $[-3, 1]$\,\si{\metre} in z-direction as this is the official evaluation range from OPV2V. 
%----------------------------
Additionally, we limit the communication range to \SI{70}{\metre}, i.e. messages from CAVs, that are not within \SI{70}{m} range of the ego are ignored.
%---------------
Furthermore, we calculate the required bandwidth for each approach at a sensor frequency of \SI{10}{\hertz}. We only consider the actual data without any communication overhead. The reported bandwidth corresponds to the average bandwidth at which each vehicle sends data to the other vehicles.
Detailed results of the conducted experiments are shown in Table \ref{tab:results}.

\begin{table}
\centering
\caption{3D average precision (AP) on the OPV2V test set together with the required bandwidth for state of the art early fusion approaches and our proposed fusion approach}
\label{tab:results}
\renewcommand{\arraystretch}{1.75}
\resizebox{\columnwidth}{!}{%
\begin{tabular}{@{}lcc@{}}
\toprule
Method                                   & AP@IoU$_{0.7}$   & \begin{tabular}[c]{@{}c@{}}Bandwidth@\SI{10}{\hertz}\\[-0.2cm]
                                                                                        \text{[}\si[per-mode=repeated-symbol]{\mega\bit\per\second}]
                                                              \end{tabular}   \\ \midrule \midrule
PV-RCNN++ No Fusion                      & 72.5              & -               \\ 
PointPillar Early Fusion                 & 80.0              & 73.1            \\ 
VoxelNet Early Fusion                    & 75.8              & 73.1            \\ 
SECOND Early Fusion                      & 81.3              & 73.1            \\ 
PIXOR Early Fusion                       & 67.8              & 73.1            \\ 
PV-RCNN++ Early Fusion                   & 79.2              & 73.1            \\
MR3D-Net No Fusion                       & 76.7              & -               \\
MR3D-Net Low Resolution                  & 82.1              & 4.3             \\
MR3D-Net Medium Resolution               & 83.2              & 8.8             \\
MR3D-Net High Resolution                 & 83.9              & 14.4            \\
MR3D-Net Mean Features                   & 81.7              & 42.2            \\
MR3D-Net Dynamic                         & 82.4              & 9.2             \\
\bottomrule
\end{tabular}
}
\end{table}

\section{Results}
\vspace*{0.2cm}
\label{sec:results}
The vehicle-local perception with PV-RCNN++ achieved an AP of \SI{72.5}{\percent}. For vehicle-local perception, no information is shared, resulting in no bandwidth requirement. 
As a collective perception baseline to compete against, we consider early fusion approaches using different object detectors. The worst result was achieved using PIXOR with \SI{67.8}{\percent}, which is a decrease of about 5 percentage points (p.p.) compared to the vehicle-local perception with PV-RCNN++. A small increase of about 3\,p.p. compared to local perception can be observed for VoxelNet. A significant increase using early fusion compared to vehicle-local perception was achieved for PointPillar and PV-RCNN++ with an AP of \SI{80.0}{\percent} and \SI{79.2}{\percent} respectively. 
%-----------------
The best early fusion result was achieved using SECOND with an AP of \SI{81.3}{\percent}, which is an increase of 8.8\,p.p. compared to vehicle-local perception with PV-RCNN++ and 1.3\,p.p. compared to PointPillar. 
%-----------------
For the average point cloud size in the OPV2V test set of \SI[per-mode=repeated-symbol]{914.9}{\kilo\byte} (see Tab.~\ref{tab:msg_sizes}) and a transmission frequency of \SI{10}{\hertz}, a bandwidth of \SI[per-mode=repeated-symbol]{73.1}{\mega\bit\per\second} per vehicle is required for all early fusion methods. 

For our MR3D-Net, the baseline for vehicle-local perception is an AP of \SI{76.7}{\percent}, which is an increase of 4.2\,p.p. compared to the PV-RCNN++ vehicle-local baseline. 

Using MR3D-Net with all collectively shared voxel grids at the low resolution of \SI[parse-numbers = false]{20\times20\times40}{\centi\meter} voxels, we achieved an AP of \SI{82.1}{\percent}. The bandwidth can be reduced to \SI[per-mode=repeated-symbol]{4.3}{\mega\bit\per\second}, which is a decrease of about \SI{94}{\percent} compared to early fusion. 
Besides the significant reduction in bandwidth requirement, MR3D-Net (low resolution) outperformed all early fusion methods in terms of AP. 

For the MR3D-Net with shared voxel grids at the medium resolution of \SI[parse-numbers = false]{10\times10\times20}{\centi\meter} voxels, we observe an increase in AP of 1.1\,p.p. to \SI{83.2}{\percent} compared to the low resolution. The bandwidth requirement is about \SI[per-mode=repeated-symbol]{9}{\mega\bit\per\second}, which is twice the bandwidth of low resolution but still a reduction of about \SI{88}{\percent} of the bandwidth for early fusion. 

MR3D-Net achieved the highest AP when all sparse voxel grids are shared at the high resolution of \SI[parse-numbers = false]{10\times10\times20}{\centi\meter} voxels. The AP of the high resolution MR3D-Net is \SI{83.9}{\percent}, which is an increase of 0.7\,p.p. compared to the medium resolution. The increase in AP is due to the higher resolution providing more detail. This increased detail results in a higher bandwidth of \SI[per-mode=repeated-symbol]{14.4}{\mega\bit\per\second}, which is still only about \SI{20}{\percent} of the bandwidth of early fusion while achieving an AP which is 2.6\,p.p. higher compared to the best early fusion method.

MR3D-Net Dynamic achieved an AP of \SI{82.4}{\percent}, which is 0.3\,p.p. higher than the low resolution and 0.8\,p.p. lower than the medium resolution while having a bandwidth requirement of \SI[per-mode=repeated-symbol]{9.2}{\mega\bit\per\second}, which is \SI[per-mode=repeated-symbol]{0.4}{\mega\bit\per\second} more than the medium resolution.
This slightly lower performance of the dynamic resolution change compared to the medium resolution may arise from the higher problem complexity induced by the dynamic resolution change.

As already mentioned, using the mean features of the points for the sparse voxel grids does not improve the performance compared to only using the center points of the voxels as features. This can be seen at the achieved result by MR3D-Net mean features, which even performs slightly worse than MR3D-Net Dynamic with an AP of \SI{81.7}{\percent}, which is a decrease in AP by 0.7\,p.p..
So transmitting only the coordinates of the sparse voxel grid is sufficient to provide a meaningful representation of the environment at these resolution levels and the mean of the point features only adds more data without providing more information about the environment.

Considering the combination of detection performance and the ability to adapt to the communication channel, MR3D-Net Dynamic overall performs the best. By using a variable resolution that dynamically adapts to the communication channel usage, the bandwidth requirement can be reduced by about \SI{87}{\percent} compared to early fusion. Regarding the detection performance, this method achieves an AP of \SI{82.4}{\percent} which is 1.1\,p.p. higher than the best early fusion approach. 

The strong performance of MR3D-Net at the lowest resolution indicates, that in cases where very limited bandwidth is available, the resolution might be reduced even further while still providing useful information to other CAVs.

Using static voxel grid resolutions also showed a significant decrease of bandwidth compared to early fusion while achieving comparable results in terms of AP. This demonstrates the advantages of the proposed sparse voxel grids as environmental representation. However, using static resolutions can lead to communication channel congestion in some cases as this methods can not dynamically adapt to the communication channel usage.

\newpage
\section{Conclusion and Future Work}
\label{sec:conclusion}

In this work, we propose a novel environment representation using sparse voxel grids together with a corresponding backbone for LiDAR-based object detectors, which allows to dynamically adapt the resolution of the data with respect to the communication channel to avoid congestion.

We demonstrated that sparse voxel grids as an environmental representation are suitable as a standardizable exchange format for sensor data for collective perception to avoid information loss during pre-processing as in late fusion approaches without overloading the communication channel as in early fusion. Our MR3D-Net fusion backbone allows to incorporate inputs with different resolutions which then are processed by sparse and submanifold convolutions at different resolution streams, that are then fused using the scatter operation.

With our MR3D-Net and the novel environment representation, we achieved a 3D average precision of \SI{82.4}{\percent} on the OPV2V dataset, significantly outperforming most early fusion methods.
In addition, we achieved a bandwidth requirement of \SI[per-mode=repeated-symbol]{9.2}{\mega\bit\per\second}, a reduction of about \SI{87}{\percent} compared to early fusion approaches.

For future research, we will evaluate further resolutions for the sparse voxel grids with respect to perception performance and bandwidth requirement. Furthermore, we will include a realistic communication channel model to investigate the channel load in more detail and create strategies for the dynamic assignment of resolutions.
Moreover, we aim to conduct evaluation on a more diverse dataset containing more variation in scenarios and number of connected and automated vehicles.

\bibliographystyle{IEEEtran}
\bibliography{literature}

\end{document}